\documentclass[runningheads]{llncs}

\usepackage{placeins}
\usepackage{times}
\usepackage[utf8]{vietnam}
\usepackage[utf8]{inputenc}
\usepackage[main=english, vietnamese]{babel}
\babeltags{vi=vietnamese}
\usepackage{type1cm}
\usepackage[T1]{fontenc}
\usepackage{microtype}
\usepackage{multirow}
\usepackage{textcomp}
\usepackage{graphicx}
\usepackage[numbers,sort&compress]{natbib}
\usepackage{marvosym}
\usepackage{float}
\usepackage{IEEEtrantools}
\usepackage{amsmath,amssymb,amsfonts}
\usepackage{pifont}
\newcommand{\cmark}{\ding{51}}%
\newcommand{\xmark}{\ding{55}}%
\usepackage{siunitx}
\usepackage{latexsym}
\usepackage{etoolbox}
\usepackage{algorithm,algpseudocode}

\usepackage[bookmarks,bookmarksopen,bookmarksdepth=3,bookmarksnumbered]{hyperref}
\hypersetup{colorlinks, citecolor=blue, linkcolor=blue, urlcolor=blue}

\makeatletter 
\renewcommand\@biblabel[1]{#1.} 
\makeatother

\begin{document}

\title{Span Labeling Approach for Vietnamese and Chinese Word Segmentation}
\titlerunning{Span Labeling Approach for Vietnamese and Chinese Word Segmentation}

\author{Duc-Vu Nguyen\inst{1,3}\and Linh-Bao Vo\inst{2,3} \and Dang Van Thin\inst{1,3} \and Ngan Luu-Thuy Nguyen\inst{2,3}}

\authorrunning{D.-V. Nguyen et al.}

\institute{Multimedia Communications Laboratory, University of Information Technology,\\Ho Chi Minh City, Vietnam \\ \email{\{vund, thindv\}@uit.edu.vn}\\ \and University of Information Technology, Ho Chi Minh City, Vietnam\\ \email{18520503@gm.uit.edu.vn, ngannlt@uit.edu.vn} \and Vietnam National University, Ho Chi Minh City, Vietnam}

\maketitle

\begin{abstract}
In this paper, we propose a span labeling approach to model n-gram information for Vietnamese word segmentation, namely \textsc{SpanSeg}. We compare the span labeling approach with the conditional random field by using encoders with the same architecture. Since Vietnamese and Chinese have similar linguistic phenomena, we evaluated the proposed method on the Vietnamese treebank benchmark dataset and five Chinese benchmark datasets. Through our experimental results, the proposed approach \textsc{SpanSeg} achieves higher performance than the sequence tagging approach with the state-of-the-art F-score of 98.31\% on the Vietnamese treebank benchmark, when they both apply the contextual pre-trained language model XLM-RoBERTa and the predicted word boundary information. Besides, we do fine-tuning experiments for the span labeling approach on BERT and ZEN pre-trained language model for Chinese with fewer parameters, faster inference time, and competitive or higher F-scores than the previous state-of-the-art approach, word segmentation with word-hood memory networks, on five Chinese benchmarks.
\end{abstract}

\section{Introduction}
Word segmentation is the first essential task for both Vietnamese and Chinese. The input of Vietnamese word segmentation (VWS) is the sequence of syllables delimited by space. In contrast, the input of Chinese word segmentation (CWS) is the sequence of characters without explicit delimiter. The use of a Vietnamese syllable is similar to a Chinese character. Despite deep learning dealing with natural language processing tasks without the word segmentation phase, the research on word segmentation is still necessary regarding the linguistic aspect. Since Vietnamese and Chinese have similar linguistic phenomena such as overlapping ambiguity in VWS \cite{hongphuong08} and in CWS \cite{sun-etal-1998-chinese}, therefore the research about VWS and CWS is a challenging problem.

Many previous approaches for VWS have been proposed. For instance, in the early stage of VWS, \citet{diend1} supposed VWS as a stochastic transduction problem. Therefore, they represented the input sentence as an unweighted Finite-State Acceptor. As a consequence, \citet{hongphuong08} proposed the ambiguity resolver using a bi-gram language model as a component in their model for VWS. After that, \citet{nguyenetal2006} used conditional random fields (CRFs) and support vector machines (SVMs) for VWS. Recently, \citet{phongnt1} utilized rules based on the predicted word boundary and threshold for the classifier in the post-processing stage to control overlapping ambiguities for VWS. Besides, \citet{datnq1} proposes a method for auto-learning rule based on the predicted word boundary for VWS. Furthermore, \citet{nguyen-2019-neural} proposed the joint neural network model for Vietnamese word segmentation, part-of-speech tagging, and dependency parsing. Lastly, \citet{nguyen-etal-2019-ws-svm} proposed feature extraction to deal with overlapping ambiguity and capturing word containing suffixes.

From our observation, the number of research and approaches for CWS is greater than VWS. The research \cite{tseng-etal-2005-conditional,song-etal-2006-chinese,chen-etal-2015-long,zhang-etal-2016-transition-based, ma-etal-2018-state, higashiyama-etal-2019-incorporating} treated CWS as a character-based sequence labeling task. The contextual feature extractions were proved helpful in CWS \cite{higashiyama-etal-2019-incorporating}. After that, neural networks were powerful for CWS \cite{chen-etal-2015-long, ma-etal-2018-state, higashiyama-etal-2019-incorporating}. The measuring word-hood for n-grams was an effective method for non-neural network model \cite{sun-etal-1998-chinese} and neural network model \cite{tian-etal-2020-improving-chinese}. Besides, the multi-criteria learning from many different datasets is a strong method \cite{chen-etal-2017-adversarial, qiu-etal-2020-concise, ke-etal-2021-pre}. Remarkably, \citet{tian-etal-2020-improving-chinese} incorporated the word-hood for n-gram into neural network model effectively.

We have an observation that most of the approaches for VWS and CWS treated word segmentation as a token-based sequence tagging problem, where the token is a syllable in VWS and character in CWS. Secondly, the intersection of VWS and CWS approaches leverages the context to model n-gram of token information, such as measuring the word-hood of the n-gram in CWS. All of the previous approaches in CWS incorporate the word-hood information as a module of their models. Therefore, our research hypothesizes whether we can model a simple model that can simulate measuring word-hood operation.

From our observation and hypothesis, we get the inspiration of span representation in constituency parsing \cite{stern-etal-2017-minimal} to propose our \textsc{SpanSeg} model for VWS and CWS. The main idea of our \textsc{SpanSeg} is to model all n-grams in the input sentence and score them. Modeling an n-gram is equivalent to find the probability of a span being a word. Via experimental results, the proposed approach \textsc{SpanSeg} achieves higher performance than the sequence tagging approach when both utilize contextual pre-trained language model XLM-RoBERTa and predicted word boundary information on the Vietnamese treebank benchmark with the state-of-the-art F-score of 98.31\%. Additionally, we do fine-tuning experiments for the span labeling method on BERT and ZEN pre-trained language model for Chinese with fewer parameters, faster inference time, and competitive or higher F-scores than the previous state-of-the-art approach, word segmentation with word-hood memory networks, on five Chinese benchmarks.

\section{The Proposed Framework}
\begin{figure}[!ht]
\centering
\includegraphics[width=\textwidth]{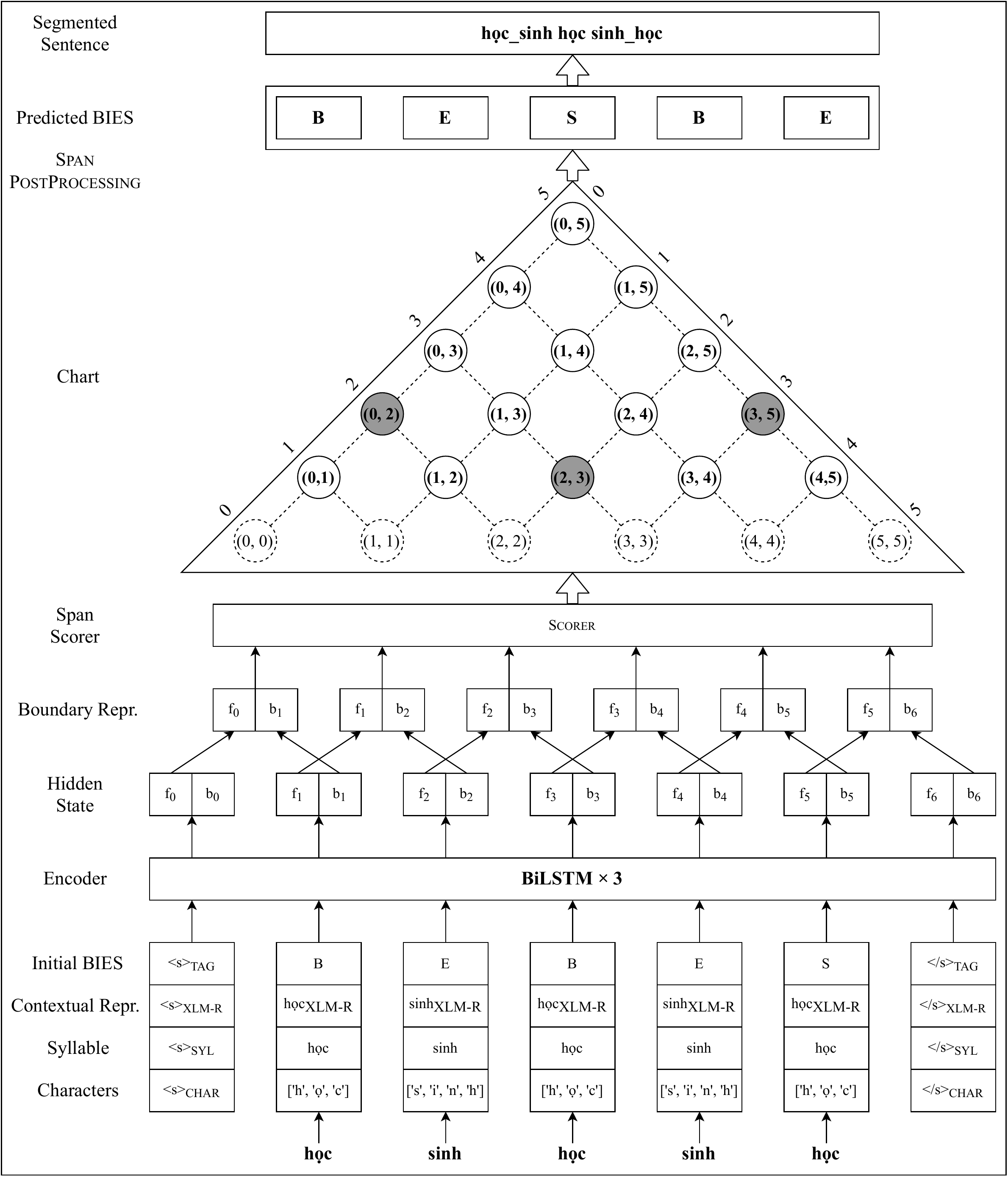}
\caption{\label{figmodel}The architecture of \textsc{SpanSeg} for VWS. The input sentence is ``\vi{học sinh học sinh học}'' (\textit{student learn biology}) including five syllables \{``\vi{học}'', ``\vi{sinh}'', ``\vi{học}'', ``\vi{sinh}'', and ``\vi{học}''\}. The gold-standard segmentation for the input sentence is ``\vi{học\_sinh học sinh\_học}'' including three words \{``\vi{học\_sinh}'', ``\vi{học}'', and ``\vi{sinh\_học}''\}. The initial BIES (Begin, Inside, End, or Singleton) word boundary tags (differing from gold-standard segmentation) were predicted by an off-the-shelf toolkit RDRsegmenter \cite{datnq1}.}
\end{figure}

Differing from previous studies, we regard word segmentation as a span labeling task. The architecture of our proposed model, namely \textsc{SpanSeg}, is illustrated in Figure~\ref{figmodel}, where the general span labeling paradigm is at the top of the figure. This paper is the first work approach to word segmentation as a span labeling task to the best of our knowledge. Before presenting the details of \textsc{SpanSeg}, we take a first look at problem representation of \textsc{SpanSeg}. In Figure~\ref{figmodel}, we consider the input sentence in the form of the index (integer type) and syllable (string type) as an array \{$0$: ``\vi{học}'', $1$: ``\vi{sinh}'', $2$: ``\vi{học}'', $3$: ``\vi{sinh}'', $4$: ``\vi{học}''\}. With this consideration, the gold-standard segmentation ``\vi{học\_sinh học sinh\_học}'' (\textit{student learn biology}) (including three words \{``\vi{học\_sinh}'', ``\vi{học}'', and ``\vi{sinh\_học}''\}) is presented by three spans $(0,2)$ (``\vi{học\_sinh}''), $(2, 3)$ (``\vi{học}''), and $(3, 5)$ (``\vi{sinh\_học}''). By approaching word segmentation as a span labeling task, we have three positive samples (three circles filled with gray color) for the input sentence in Figure~\ref{figmodel}, whereas other circles filled with white color with solid border are negative samples for the input sentence in Figure~\ref{figmodel}. Also, in Figure~\ref{figmodel}, we note that all circles with dashed border (e.g, spans $(0,0)$, $(1,1)$, $\dots$, $(n,n)$, where $n$ is the length of the input sentence) are skipped in \textsc{SpanSeg} because they do not represent spans.

After presenting \textsc{SpanSeg}, in the rest of this section, we firstly introduce problem representation of word segmentation as a span labeling task (in subsection~\ref{subproblem}). Secondly, we introduce the proposed span post-processing algorithm for word segmentation (in subsection~\ref{spanpost}). The first and second subsections are two important points of our research. Thirdly, we describe the span scoring module (in subsection~\ref{scoringmodule}). In the last two subsections, we provide the architecture encoder for VWS and CWS. We describe the model \textsc{SpanSeg} for VWS (in subsection~\ref{spansegvi}). Lastly, we describe \textsc{SpanSeg} for CWS (in subsection~\ref{spansegzh}).

\subsection{Word segmentation as span labeling task for Vietnamese and Chinese}
\label{subproblem}
The input sentence of word segmentation task is a sequence of tokens $\mathcal{X} = x_1 x_2 \dots x_{n}$ with the length of $n$. The token $x_i$ is a syllable or character toward Vietnamese or Chinese, respectively. Given the input $\mathcal{X}$, the output of word segmentation is a sequence of words $\mathcal{W} = w_1 w_2 \dots w_{m}$ with the length of $m$, where $1 \leq m \leq n$. We have a property that the word $w_j$ is constituted by one token or consecutive tokens. So, we use the sequence of tokens $x_i x_{i+1} \dots x_{i+k-1}$ for denoting the word $w_j$ be constituted by $k$ consecutive tokens beginning at token $x_i$, where $1 \leq k \leq n$ (concretely, $k = 1$ representing single words and $2 \leq k \leq n$ representing compound words for both Vietnamese and Chinese). Inspired by the work of \citet{stern-etal-2017-minimal} for constituency parsing, we use the span $(i - 1, i - 1 + k)$ to represent the word constituted by $k$ consecutive tokens $x_i x_{i+1} \dots x_{i+k-1}$ beginning at token $x_i$. Therefore, the goal of the span labeling task for both VWS and CWS is to find the list of spans $\hat{\mathcal{S}}$ such that every token $x_i$ is spanned, and there is no overlapping between every two spans. Formally, the word segmentation model as span labeling task for both VWS and CWS can be formalized as:
\begin{IEEEeqnarray}{rCl}
\hat{\mathcal{S}} = \textsc{SpanPostProcessing}(\hat{\mathcal{Y}})
\end{IEEEeqnarray}
where \textsc{SpanPostProcessing}($\cdot$) simply is a algorithm for producing the word segmentation boundary satisfying non-overlapping between every two spans. The $\hat{\mathcal{Y}}$ is the set of predicted spans as following:
\begin{IEEEeqnarray}{rCl}
\hat{\mathcal{Y}} = \{(l, r) | 0 \leq l \leq n - 1 ~\text{and}~l < r \leq n~\text{and}~\textsc{Scorer}(\mathcal{X}, l, r) > 0.5\}
\end{IEEEeqnarray}
where $n$ is the length of the input sentence. The \textsc{Scorer}($\cdot$) is the scoring module for the span $(l, r)$ of sentence $\mathcal{X}$. The output of \textsc{Scorer}($\cdot$) has a value in the range of 0 to 1. In our research, we choose the sigmoid function as an activation function at the last layer of \textsc{Scorer}($\cdot$) module. Lastly, the word segmentation as a span labeling task is the binary classification problem. We use the binary cross-entropy loss for the cost function as following:
\begin{align}
J(\theta) = -\frac{1}{|\mathcal{D}|}\sum_{\mathcal{X, S} \in \mathcal{D}} \biggl(\frac{1}{(n(n+1))/2}&\sum_{l=0}^{n-1} \sum_{r=l+1}^{n} \left[(l,r) \in \mathcal{S}\right]\log\big(\textsc{Scorer}(\mathcal{X}, l, r)\big) \nonumber\\
{}+{}&\left[(l,r) \notin \mathcal{S}\right]\log\big(1 - \textsc{Scorer}(\mathcal{X}, l, r)\big) \biggr)
\end{align}
where $\mathcal{D}$ is the training set and $|\mathcal{D}|$ is the size of training set. For each pair ($\mathcal{X, S}$) in training set $\mathcal{D}$, we compute binary cross-entropy loss for all spans $(l, r)$, where $0 \leq l \leq n - 1 ~\text{and}~l < r \leq n$, and $n$ is the length of sentence $\mathcal{X}$. The term $\left[(l,r) \in \mathcal{S}\right]$ has the value of $1$ if span $(l, r)$ belongs to the list $\mathcal{S}$ of sentence $\mathcal{X}$ and conversely, of 0. Similarly, the term $\left[(l,r) \notin \mathcal{S}\right]$ has the value of $1$ if span $(l, r)$ does not belong to the list $\mathcal{S}$ of sentence $\mathcal{X}$ and conversely, of 0. Lastly, we make a note that in our training and prediction progress, we will discard spans with length greater than 7 for both Vietnamese and Chinese (7 is maximum n-gram length following \cite{diao-etal-2020-zen} for Chinese, so we decide to choose 7 for Vietnamese according to the statistics in the work of \cite{nguyen-etal-2019-ws-svm}).

\subsection{Post-Processing Algorithm For Predicted Spans}
\label{spanpost}

In the previous subsection~\ref{subproblem}, we presented word segmentation as a span labeling task for Vietnamese and Chinese. In this subsection, we present our proposed post-processing algorithm for predicted spans from the span labeling problem. However, we found that in the predicted spans set $\hat{\mathcal{Y}}$ there exists overlapping between some two spans. We deal with the overlapping ambiguity by choosing the spans with the highest score and removing the rest. The overlapping ambiguity phenomenon occurs when our \textsc{SpanSeg} predicts compound words. It occurs in our \textsc{SpanSeg} and other word segmenters on Vietnamese \cite{hongphuong08} and Chinese \cite{sun-etal-1998-chinese}.

Apart from overlapping ambiguity, our \textsc{SpanSeg} faces the missing word boundary problem. That problem can be caused by originally predicted spans or as a result of solving overlapping ambiguity. We choose the missing word boundary based on all predicted spans $(i - 1, i - 1 + k)$ with $k = 1$ for single words to deal with the missing word boundary problem. To sum up, our proposed post-processing algorithm for predicted spans from the span labeling problem, namely \textsc{SpanPostProcessing}, deals with overlapping ambiguity and missing spans from predicted spans. The detail of our \textsc{SpanPostProcessing} is presented in Algorithm~\ref{alg}.

\begin{algorithm}[!ht]
  \caption{\textsc{SpanPostProcessing}}\label{alg}
  \begin{algorithmic}[1]
    \Require
      \Statex The input sentence $\mathcal{X}$ with the length of $n$;
      \Statex The scoring module \textsc{Scorer}($\cdot$) for any span $(l, r)$ in $\mathcal{X}$, where $0 \leq l \leq n - 1 ~\text{and}~l < r \leq n$;
      \Statex The set of predicted spans $\hat{\mathcal{Y}}$, sorted in ascending order.
    \Ensure
      \Statex The list of valid predicted spans $\hat{\mathcal{S}}$, satisfying non-overlapping between every two spans.
    \Statex
    \State $\hat{\mathcal{S}}_\text{novlp} = [(0, 0)]$ \Comment{The list of predicted spans without overlapping ambiguity.}
    \State $\hat{\mathcal{S}} = []$ \Comment{The final list of valid predicted spans.}
    \For{$\hat{y}$ \textbf{in} $\hat{\mathcal{Y}}$} \Comment{The $\hat{y}[0]$ is the left boundary and $\hat{y}[1]$ is the right boundary of each span $\hat{y}$.}
        \If{$\hat{\mathcal{S}}_\text{novlp}[\text{-}1][1] < \hat{y}[0]$} \Comment{Check for missing boundary.}
        \State $\hat{\mathcal{S}}_\text{novlp}$.\textbf{append}$\big((\hat{\mathcal{S}}_\text{novlp}[\text{-}1][1], \hat{y}[0])\big)$ \Comment{Add the missing span to $\hat{\mathcal{S}}_\text{novlp}$}
        \EndIf
        \If{$\hat{\mathcal{S}}_\text{novlp}[\text{-}1][0] \leq \hat{y}[0] < \hat{\mathcal{S}}_\text{novlp}[\text{-}1][1]$}  \Comment{Check for overlapping ambiguity.}
        \If{$\textsc{Scorer}(\mathcal{X}, \hat{\mathcal{S}}_\text{novlp}[\text{-}1][0], \hat{\mathcal{S}}_\text{novlp}[\text{-}1][1]) < \textsc{Scorer}(\mathcal{X}, \hat{y}[0], \hat{y}[1])$}
            \State $\hat{\mathcal{S}}_\text{novlp}$.\textbf{pop}()  \Comment{Remove the span causing overlapping with the lower score than $\hat{y}$.}
            \State $\hat{\mathcal{S}}_\text{novlp}$.\textbf{append}$\big((\hat{y}[0], \hat{y}[1])\big)$ \Comment{Add the span $\hat{y}$ to $\hat{\mathcal{S}}_\text{novlp}$.}
        \EndIf
        \Else{}
        \State $\hat{\mathcal{S}}_\text{novlp}$.\textbf{append}$\big((\hat{y}[0], \hat{y}[1])\big)$ \Comment{Add the span $\hat{y}$ to $\hat{\mathcal{S}}_\text{novlp}$.}
        \EndIf
    \EndFor
    \If{$\hat{\mathcal{S}}_\text{novlp}[\text{-}1][1] < n$} \Comment{Check for missing boundary.}   
        \State $\hat{\mathcal{S}}_\text{novlp}$.\textbf{append}$\big((\hat{\mathcal{S}}_\text{novlp}[\text{-}1][1], n)\big)$ \Comment{Add the missing span to $\hat{\mathcal{S}}_\text{novlp}$}
    \EndIf
    \For{$i, \hat{y}$ \textbf{in} \textbf{enumerate}($\hat{\mathcal{S}}_\text{novlp}$)} \Comment{The $\hat{y}[0]$ is the left boundary and $\hat{y}[1]$ is the right boundary of each span $\hat{y}$, and $i$ is the index of $\hat{y}$ in list $\hat{\mathcal{S}}_\text{novlp}$.}
      \If{$0 < i$ \textbf{and} $\hat{\mathcal{S}}_\text{novlp}[i-1][1] < \hat{y}[0]$} \Comment{Check for missing boundary.}
      \State \textit{missed\_boundaries} $= \big[\hat{\mathcal{S}}_\text{novlp}[i-1][1]\big]$
        \For{\textit{bound} \textbf{in} \textbf{range}$\big(\hat{\mathcal{S}}_\text{novlp}[i-1][1], \hat{y}[0]\big)$}
          \If{$\textsc{Scorer}(\mathcal{X}, \textit{bound}, \textit{bound} + 1) > 0.5$}\Comment{Check for single word.}
            \State \textit{missed\_boundaries}.\textbf{append}($\textit{bound} + 1$)
          \EndIf
        \EndFor
        \State \textit{missed\_boundaries}.\textbf{append}($\hat{y}[0]$)
        \For{\textit{j} \textbf{in} \textbf{range}$\big(\textbf{len}(\textit{missed\_boundaries}) - 1\big)$}
          \State $\hat{\mathcal{S}}$.\textbf{append}$\big((\textit{missed\_boundaries}[j], \textit{missed\_boundaries}[j+1])\big)$\Comment{Add the missing span to $\hat{\mathcal{S}}$}
        \EndFor
      \EndIf
      \State $\hat{\mathcal{S}}$.\textbf{append}$\big(\hat{y}[0], \hat{y}[1]\big)$ \Comment{Add the non-overlapping span to $\hat{\mathcal{S}}$}
    \EndFor
  \end{algorithmic}
\end{algorithm}

\subsection{Span Scoring Module}
\label{scoringmodule}
In two previous subsections~\ref{subproblem} and ~\ref{spanpost}, we presented two critical points of our research. There we mentioned the \textsc{Scorer}($\cdot$) module many times. In this section, we present \textsc{Scorer}($\cdot$) module. It is based on the familiar module that name Biaffine \cite{dozat2017deep}. While \citet{ijcai2020-560} experimenting with the Biaffine module for constituency parsing, we use the Biaffine module for span labeling word segmentation. The Biaffine module is used in \cite{dozat2017deep} to capture the directed relation between two words in a sentence for dependency parsing. In the constituency parsing problem, \citet{ijcai2020-560} used the Biaffine module to find the representation of phrases. Our research uses the Biaffine module to model the representation of n-gram for the word segmentation task.

As we can see in Figure~\ref{figmodel}, each token $x_i$ in the input sentence has two context-aware word representations including left and right boundary representations except the begin (``<s>'') and end (``</s>'') tokens. In case we use the BiLSTM (Bidirectional Long Short Term Memory) encoder, the left boundary representation of token $x_i$ is the concatenation of the hidden state forward vector $\textbf{f}_{i-1}$ and the hidden state backward vector $\textbf{b}_{i}$ and the right boundary representation of token $x_i$ is the concatenation of the hidden state forward vector $\textbf{f}_{i}$ and the hidden state backward vector $\textbf{b}_{i+1}$, following \citet{stern-etal-2017-minimal}. In case we use BERT \cite{devlin-etal-2019-bert} or ZEN \cite{diao-etal-2020-zen} encoder, we chunk the last hidden state vector into two vectors with the same size as forward and backward vectors of the BiLSTM encoder. Even though we use the BiLSTM, BERT, or ZEN encoder, we always have the left and right boundary representation for each token $x_i$ in the input sentence. Therefore, in Figure~\ref{figmodel}, we see that the right boundary representation $\textbf{f}_{i}\oplus \textbf{b}_{i+1}$ of token $x_i$ is the left boundary representation of token $\textbf{x}_{i+1}$. As the work of \citet{ijcai2020-560}, we use two MLPs to make the difference between the right boundary representation of token $x_i$ and the left boundary representation of token $x_{i+1}$. To sum up, we have the left $\textbf{r}_i^{\text{left}}$ and right $\textbf{r}_i^{\text{right}}$ boundary representations of token $x_{i}$ as following:
\begin{IEEEeqnarray}{rCl}
\textbf{r}_i^{\text{left}} &=& \text{MLP}^{\text{left}}(\textbf{f}_{i-1}\oplus \textbf{b}_{i})\\
\textbf{r}_i^{\text{right}} &=& \text{MLP}^{\text{right}}(\textbf{f}_{i}\oplus \textbf{b}_{i+1})
\end{IEEEeqnarray}
Finally, inspired by \citet{ijcai2020-560}, given the input sentence $\mathcal{X}$, the span scoring module \textsc{Scorer}($\cdot$) for span $(l, r)$ in our \textsc{SpanSeg} model is computed by using a biaffine operation over the left boundary representation of token $x_l$ and the right boundary representation of token $x_r$ as following:
\begin{IEEEeqnarray}{rCl}
\textsc{Scorer}(\mathcal{X}, l, r) = \text{sigmoid}\bigg( \begin{bmatrix}\textbf{r}_l^{\text{left}} \\ 1\end{bmatrix}^{\text{T}}\textbf{W}\textbf{r}_r^{\text{right}}\bigg)
\end{IEEEeqnarray}
where $\textbf{W} \in \mathbb{R}^{d\times d}$. To sum up, the \textsc{Scorer}($\mathcal{X}, l, r$) gives us a score to predict whether a span $(l, r)$ is a word.

\subsection{Encoder and input representation for VWS}
\label{spansegvi}
\begin{sloppypar}In three previous subsection~\ref{subproblem}, \ref{spanpost}, and \ref{scoringmodule}, we describe three mutual parts of the \textsc{SpanSeg} model for Vietnamese and Chinese. In this subsection, we present the encoder and the input representation for VWS of the \textsc{SpanSeg} model. Firstly, the default configuration of \textsc{SpanSeg} for the input representation of token $x_i$ is composed as following:
\begin{align}
\textbf{default\_embedding}_i =  \big(&\textbf{static\_syl\_embedding}_{i} \nonumber\\
& + \textbf{dynamic\_syl\_embedding}_{i}\big) \oplus \textbf{char\_embedding}_i
\end{align}
where the symbol $\oplus$ denotes the concatenation operation. The $\textbf{static\_syl\_embedding}_{i}$ is extracted from the pre-trained Vietnamese syllable embedding with the dimension of 100 provided by \citet{nguyen-etal-2017-word}. So, the dimension of vector $\textbf{dynamic\_syl\_embedding}_{i}$ also is 100. We initialize randomly and update the value of $\textbf{dynamic\_syl\_embedding}_{i}$ in the training progress. We do not update the value of $\textbf{static\_syl\_embedding}_{i}$ during training model. Besides, we also use a character embedding for the input representation by using BiLSTM network for sequence of characters in token $x_i$ to obtain $\textbf{char\_embedding}_i$.\end{sloppypar}

The default configuration does not utilize the Vietnamese predicted word boundary information as many previous works on VWS did. Following the work of \citet{nguyen-2019-neural}, we additionally use the boundary BIES tag embedding for the input representation of token $x_i$. Therefore, the second configuration of \textsc{SpanSeg}, namely \textsc{SpanSeg} (TAG) is presented as following:
\begin{align}
\textbf{default\_tag\_embedding}_i = \textbf{default\_embedding}_i \oplus \textbf{bies\_tag\_embedding}_i
\end{align}
where the value of $\textbf{bies\_tag\_embedding}_i$ (with the dimension of 100) is initialized randomly and updated; and the boundary BIES tag is predicted by the off-the-shelf toolkit RDRsegmenter \cite{datnq1}.

Recently, many contextual pre-trained language models were proposed inspired by the work of \citet{devlin-etal-2019-bert}. However, our research utilizes contextual pre-trained multilingual language model XLM-Roberta (XLM-R) \cite{conneau-etal-2020-unsupervised} with the \textit{base} architecture for VWS since there is no contextual pre-trained monolingual language model for Vietnamese at this time. So, the third configuration of \textsc{SpanSeg}, namely \textsc{SpanSeg} (XLM-R), is presented as following:
\begin{align}
\textbf{default\_xlmr\_embedding}_i = \textbf{default\_embedding}_i \oplus \textbf{xlmr\_embedding}_i
\end{align}
where the $\textbf{xlmr\_embedding}_i$ is the projected vector from the hidden state of \textit{the last four layers} of the XLM-R model. The dimension of $\textbf{xlmr\_embedding}_i$ is 100. We do not update parameters of the XLM-R model during the training process.

Lastly, we make the fourth configuration for \textsc{SpanSeg}, namely \textsc{SpanSeg} (TAG + XLM-R). This configuration aims to combine all syllables, characters, predicted word boundaries, and contextual information for VWS.
\begin{align}
\textbf{default\_tag\_xlmr\_embedding}_i =~& \textbf{default\_embedding}_i \nonumber\\
 & \oplus \textbf{bies\_tag\_embedding}_i \oplus \textbf{xlmr\_embedding}_i
\end{align}

After we have the input representation for each token $x_i$ of the input sentence $\mathcal{X}$, we feed them into the BiLSTM network to obtain the forward $f_i$ and backward $b_i$ vectors. The forward $f_i$ and backward $b_i$ vectors is used in the \textsc{Scorer}($\cdot$) module in subsection~\ref{scoringmodule}.

\subsection{Encoder and input representation for CWS}
\label{spansegzh}
To make a fair comparison to the state-of-the-art model for CWS, we used the same encoder as the work of \citet{tian-etal-2020-improving-chinese}. Following the work \cite{tian-etal-2020-improving-chinese}, we choose two BERT \cite{devlin-etal-2019-bert} and ZEN \cite{diao-etal-2020-zen} encoders with the \textit{base} architecture. The BERT and ZEN are two famous encoders utilizing contextual information for Chinese language processing, in which the ZEN encoder enhances n-gram of characters information. For each character $x_i$ in the input sentence $\mathcal{X}$, we chunk the hidden state vector of \textit{the last layer} of BERT or ZEN into two vectors with the same size as the forward $f_i$ and backward $b_i$ vectors in the BiLSTM network. Finally, the forward $f_i$ and backward $b_i$ vectors are used in the \textsc{Scorer}($\cdot$) module in subsection~\ref{scoringmodule}. We update the parameters of BERT and ZEN in training progress following the work of \citet{tian-etal-2020-improving-chinese}.

\section{Experimental Settings}
\subsection{Datasets}

The largest VWS benchmark dataset\footnote{The details of VTB dataset are presented at \url{https://vlsp.org.vn/vlsp2013/eval/ws-pos}.} is a part of the Vietnamese treebank (VTB) project \cite{nguyen-etal-2009-building}. We use the same split as the work of \citet{datnq1}. The summary of the VTB dataset for the word segmentation task is provided in Table~\ref{vtbstat}.

\begin{table}[ht]
\caption{\label{vtbstat}Statistics of the Vietnamese treebank dataset for word segmentation. We provide the number of sentences, characters, syllables, words, character types, syllable types, word types. We also compute the out-of-vocabulary (OOV) rate as the percentage of unseen words in the development and test set.}
\centering
\resizebox{0.535\textwidth}{!}{%
\begin{tabular}{l|r|r|r}
\hline
\multirow{2}{*}{} & \multicolumn{3}{c}{\textbf{VTB}} \\ \cline{2-4} 
 & \multicolumn{1}{c|}{Train} & \multicolumn{1}{c|}{Dev} & \multicolumn{1}{c}{Test} \\ \hline
\# sentences & 74,889 & 500 & 2,120 \\ \hline
\# characters & 6,779,116 & 55,476 & 307,932 \\ \hline
\# syllables & 2,176,398 & 17,429 & 96,560 \\ \hline
\# words & 1,722,271 & 13,165 & 66,346 \\ \hline
\# character types & 155 & 117 & 121 \\ \hline
\# syllable types & 17,840 & 1,785 & 2,025 \\ \hline
\# word types & 41,355 & 2,227 & 3,730 \\ \hline
OOV Rate & - & 2.2 & 1.6 \\ \hline
\end{tabular}%
}
\end{table}

For evaluating our \textsc{SpanSeg} on CWS, we employ five benchmark datasets including MSR, PKU, AS, CityU (from SIGHAN 2005 Bakeoff \cite{emerson-2005-second}), and CTB6 \cite{xue_xia_chiou_palmer_2005}. We convert traditional Chinese characters in AS, and CityU into simplified ones following previous studies \cite{chen-etal-2015-long, qiu-etal-2020-concise, tian-etal-2020-improving-chinese}. We follow the official training/test data split of MSR, PKU, AS, and CityU, in which we randomly extract 10\% of the training dataset for development as many previous works. For CTB6, we the same split as the work of \citet{tian-etal-2020-improving-chinese}. For pre-processing phase of all CWS dataset in our research, we inherit the process\footnote{\url{https://github.com/SVAIGBA/WMSeg}} of \citet{tian-etal-2020-improving-chinese}. The summary of five Chinese benchmark datasets for the word segmentation task is presented in Table~\ref{zhstat}.

\begin{table}[H]
\caption{\label{zhstat}Statistics of five Chinese benchmark dataset for word segmentation. We provide the number of sentences, characters, words, character types, word types. We also compute the out-of-vocabulary (OOV) rate as the percentage of unseen words in the test set.}
\resizebox{\textwidth}{!}{%
\begin{tabular}{l|r|r|r|r|r|r|r|r|r|r|r}
\hline
\multirow{2}{*}{} & \multicolumn{2}{c|}{\textbf{MSR}} & \multicolumn{2}{c|}{\textbf{PKU}} & \multicolumn{2}{c|}{\textbf{AS}} & \multicolumn{2}{c|}{\textbf{\textsc{CityU}}} & \multicolumn{3}{c}{\textbf{CTB6}} \\ \cline{2-12} 
 & \multicolumn{1}{c|}{Train} & \multicolumn{1}{c|}{Test} & \multicolumn{1}{c|}{Train} & \multicolumn{1}{c|}{Test} & \multicolumn{1}{c|}{Train} & \multicolumn{1}{c|}{Test} & \multicolumn{1}{c|}{Train} & \multicolumn{1}{c|}{Test} & \multicolumn{1}{c|}{Train} & \multicolumn{1}{c|}{Dev} & \multicolumn{1}{c}{Test} \\ \hline
\# sentences & 86,918 & 3,985 & 19,054 & 1,944 & 708,953 & 14,429 & 53,019 & 1,492 & 23,420 & 2,079 & 2,796 \\ \hline
\# characters & 4,050,469 & 184,355 & 1,826,448 & 172,733 & 8,368,050 & 197,681 & 2,403,354 & 67,689 & 1,055,583 & 100,316 & 134,149 \\ \hline
\# words & 2,368,391 & 106,873 & 1,109,947 & 104,372 & 5,449,581 & 122,610 & 1,455,630 & 40,936 & 641,368 & 59,955 & 81,578 \\ \hline
\# character types & 5,140 & 2,838 & 4,675 & 2,918 & 5,948 & 3,578 & 4,806 & 2,642 & 4,243 & 2,648 & 2,917 \\ \hline
\# word types & 88,104 & 12,923 & 55,303 & 13,148 & 140,009 & 18,757 & 68,928 & 8,989 & 42,246 & 9,811 & 12,278 \\ \hline
OOV Rate & - & 2.7 & - & 5.8 & - & 4.3 & - & 7.2 & - & 5.4 & 5.6 \\ \hline
\end{tabular}%
}
\end{table}

\subsection{Model Implementation}
\subsubsection{The detail of \textsc{SpanSeg} for Vietnamese}
For the encoder mentioned in the subsection~\ref{spansegvi}, the number of layers of BiLSTM is $3$, and the hidden size of BiLSTM is 400. The size of MLPs mentioned in the subsection~\ref{scoringmodule} is 500. The dropout rate for embedding, BiLSTM, and MLPs is 0.33. We inherit hyper-parameters from the work of \cite{dozat2017deep}. We trained all models up to 100 with the early stopping strategy with patience epochs of 20. We used AdamW optimizer \cite{loshchilov2019decoupled} with the default configuration and learning rate of $\text{10}^{\text{-3}}$. The batch size for training and evaluating is up to 5000.

\subsubsection{The detail of \textsc{SpanSeg} for Chinese}
For the encoder mentioned in the subsection~\ref{spansegzh}, we do fine-tuning experiments based on BERT \cite{devlin-etal-2019-bert} and ZEN \cite{diao-etal-2020-zen} encoders. The size of MLPs mentioned in the subsection~\ref{scoringmodule} is 500. The dropout rate for BERT and ZEN is 0.1. We trained all models up to 30 with the early stopping strategy with patience epochs of 5. We used AdamW optimizer \cite{loshchilov2019decoupled} with the default configuration and learning rate of $\text{10}^{\text{-5}}$. The batch size for training and evaluating is 16.

\section{Results and Analysis}

\subsection{Main Results}

For VWS, we also implement the BiLSTM-CRF model with the same backbone and hyper-parameters as our \textsc{SpanSeg}. The overall results are presented in Table~\ref{mainvtb}. On the default configuration, our \textsc{SpanSeg} gives a higher result than BiLSTM-CRF with the F-score of 97.76\%. On the configuration with pre-trained XLM-R, our \textsc{SpanSeg} (XLM-R) gives a higher result than BiLSTM-CRF (XLM-R) with the F-score of 97.95\%. On the configuration with predicted boundary BIES tag from off-the-shelf toolkit RDRsegmenter \cite{datnq1}, the BiLSTM-CRF (TAG) gives a higher result than our \textsc{SpanSeg} (TAG) with the F-score of 98.10\%. Finally, on the configuration with a combination of all features, our \textsc{SpanSeg} (TAG+XLM-R) gives a higher result than BiLSTM-CRF (TAG+XLM-R) with the F-score of 98.31\%, which is also the state-of-the-art performance on VTB. We can see that the contextual information is essential for \textsc{SpanSeg} since \textsc{SpanSeg} models the left and right boundary of a word rather than the between to consecutive tokens.

\begin{table}[ht]
\caption{\label{mainvtb}Performance (F-score) comparison between \textsc{SpanSeg} (with different configurations) and previous state-of-the-art models on the test set of VTB dataset.}
\centering
\resizebox{0.65\textwidth}{!}{%
\begin{tabular}{l|c|c|c|c}
\hline
\multirow{2}{*}{} & \multicolumn{4}{c}{\textbf{VTB}} \\ \cline{2-5} 
 & P & R & F & $\text{R}_\text{OOV}$ \\ \hline
vnTokenizer \cite{hongphuong08} & 96.98 &  97.69 &  97.33 & - \\ \hline
JVnSegmenter-Maxent \cite{nguyenetal2006} & 96.60 &  97.40 &  97.00 & -\\ \hline
JVnSegmenter-CRFs \cite{nguyenetal2006} &  96.63 &  97.49 & 97.06 & - \\ \hline
DongDu \cite{luu2012} & 96.35 & 97.46 & 96.90 & - \\ \hline
UETsegmenter \cite{phongnt1}  & 97.51 & 98.23 & 97.87 & - \\ \hline
RDRsegmenter \cite{datnq1} & 97.46 & 98.35 & 97.90 & - \\ \hline
UITsegmenter \cite{nguyen-etal-2019-ws-svm} & 97.81 & \textbf{98.57} & 98.19 & - \\ \hline
BiLSTM-CRF & 97.42 & 97.84 & 97.63 & 72.47 \\ 
\textsc{SpanSeg} & 97.58 & 97.94 & 97.76 & \textbf{74.65} \\ 
BiLSTM-CRF (XLM-R) & 97.69 & 97.99 & 97.84 & 72.66 \\ 
\textsc{SpanSeg} (XLM-R) & 97.75 & 98.16 & 97.95 & 70.01 \\ 
BiLSTM-CRF (TAG) & 97.91 & 98.28 & 98.10 & 69.16 \\ 
\textsc{SpanSeg} (TAG) & 97.67 & 98.28 & 97.97 & 65.94 \\ 
BiLSTM-CRF (TAG+XLM-R) & 97.94 & 98.44 & 98.19 & 68.87 \\ 
\textsc{SpanSeg} (TAG+XLM-R) & \textbf{98.21} & 98.41 & \textbf{98.31} & 72.28 \\ \hline
\end{tabular}%
}
\end{table}

For CWS, we presented the performances of our \textsc{SpanSeg} in Table~\ref{zhmain}. We do not compare our method with previous studies approaching multi-criteria learning since simply the training data is different. Our research focuses on the comparison between our \textsc{SpanSeg} and sequence tagging approaches. Firstly, we can see that our \textsc{SpanSeg} (BERT) achieves higher results than state-of-the-art methods \textsc{WMSeg} (BERT-CRF) \cite{tian-etal-2020-improving-chinese} on four datasets including MSR (98.31\%), PKU (96.56\%), AS (96.62\%), and CTB6 (97.26\%) except CityU (97.74\%). Our \textsc{SpanSeg} (ZEN) do not achieve the stable performance as \textsc{SpanSeg} (BERT). The potential reason for this problem is that both ZEN \cite{diao-etal-2020-zen} encoder and our \textsc{SpanSeg} try to model n-gram of Chinese characters causing inconsistency.

\begin{table}[ht]
\caption{\label{zhmain}Performance (F-score) comparison between \textsc{SpanSeg} (BERT and ZEN) and previous state-of-the-art models on the test set of five Chinese benchmark datasets. The symbol [$\bigstar$] denotes the methods learning from data annotated through different segmentation criteria, which means that the labeled training data are different from the rest.}
\resizebox{\textwidth}{!}{%
\begin{tabular}{l|c|c|c|c|c|c|c|c|c|c}
\hline
\multirow{2}{*}{} & \multicolumn{2}{c|}{\textbf{MSR}} & \multicolumn{2}{c|}{\textbf{PKU}} & \multicolumn{2}{c|}{\textbf{AS}} & \multicolumn{2}{c|}{\textbf{\textsc{CityU}}} & \multicolumn{2}{c}{\textbf{CTB6}} \\ \cline{2-11} 
 & F & $\text{R}_\text{OOV}$ & F & $\text{R}_\text{OOV}$ & F & $\text{R}_\text{OOV}$ & F & $\text{R}_\text{OOV}$ & F & $\text{R}_\text{OOV}$ \\ \hline
\citet{chen-etal-2015-long} & 97.40 & - & 96.50 & - & - & - & - & - & 96.00 & - \\
\citet{xu-sun-2016-dependency} & 96.30 & - & 96.10 & - & - & - & - & - & 95.80 & - \\
\citet{zhang-etal-2016-transition-based} & 97.70 & - & 95.70 & - & - & - & - & - & 95.95 & - \\
\citet{chen-etal-2017-adversarial} [$\bigstar$] & 96.04 & 71.60 & 94.32 & 72.64 & 94.75 & 75.34 & 95.55 & 81.40 & - & - \\
\citet{wang-xu-2017-convolutional} & 98.00 & - & 96.50 & - & - & - & - & - & - & - \\
\citet{zhou-etal-2017-word} & 97.80 & - & 96.00 & - & - & - & - & - & 96.20 & - \\
\citet{ma-etal-2018-state} & 98.10 & 80.00 & 96.10 & 78.80 & 96.20 & 70.70 & 97.20 & 87.50 & 96.70 & 85.40 \\
\citet{GongCGQ19} & 97.78 & 64.20 & 96.15 & 69.88 & 95.22 & 77.33 & 96.22 & 73.58 & - & - \\
\citet{higashiyama-etal-2019-incorporating} & 97.80 & - & - & - & - & - & - & - & 96.40 & - \\
\citet{qiu-etal-2020-concise} [$\bigstar$] & 98.05 & 78.92 & 96.41 & 78.91 & 96.44 & 76.39 & 96.91 & 86.91 & - & - \\\hline
\textsc{WMSeg (BERT-CRF)} \cite{tian-etal-2020-improving-chinese} & 98.28 & \textbf{86.67} & 96.51 & \textbf{86.76} & 96.58 & 78.48 & 97.80 & 87.57 & 97.16 & 88.00 \\
\textsc{WMSeg (ZEN-CRF)} \cite{tian-etal-2020-improving-chinese} & \textbf{98.40} & 84.87 & 96.53 & 85.36 & \textbf{96.62} & \textbf{79.64} & 97.93 & \textbf{90.15} & \textbf{97.25} & \textbf{88.46} \\ \hline
\textsc{MetaSeg} \cite{ke-etal-2021-pre} [$\bigstar$] & 98.50 & - & 96.92 & - & 97.01 & - & 98.20 & - & 97.89 & - \\ \hline
\textsc{SpanSeg (BERT)} & 98.31 & 85.32 & \textbf{96.56} & 85.53 & \textbf{96.62} & 79.36 & 97.74 & 87.45 & \textbf{97.25} & 87.91 \\
\textsc{SpanSeg (ZEN)} & 98.35 & 85.66 & 96.35 & 83.66 & 96.52 & 78.43 & \textbf{97.96} & 90.11 & 97.17 & 87.76 \\
\hline
\end{tabular}%
}
\end{table}

\begin{table}[ht]
\caption{\label{timetab}Statistics of model size (MB) and inference time (minute) of \textsc{WMSeg} \cite{tian-etal-2020-improving-chinese} and our \textsc{SpanSeg} dealing with the training set of the AS dataset on Chinese. We use the same batch size as the work of \citet{tian-etal-2020-improving-chinese}. The inference time is done by using Tesla P100-PCIE GPU with memory size of 16,280 MiB via Google Colaboratory.}
\centering
\resizebox{0.65\textwidth}{!}{%
\begin{tabular}{l|c|r|c|r}
\hline
\multicolumn{1}{r|}{\multirow{2}{*}{}} & \multicolumn{2}{c|}{\textbf{BERT Encoder}} & \multicolumn{2}{c}{\textbf{ZEN Encoder}} \\ \cline{2-5} 
\multicolumn{1}{r|}{} & \textbf{\textsc{WMSeg}} & \multicolumn{1}{c|}{\textbf{\textsc{SpanSeg}}} & \textbf{\textsc{WMSeg}} & \multicolumn{1}{c}{\textbf{\textsc{SpanSeg}}} \\ \hline
Size (MB) & \multicolumn{1}{r|}{704} & 397 & \multicolumn{1}{r|}{1,150} & 872 \\ \hline
Inference Time (minute) & \multicolumn{1}{r|}{28} & 15 & \multicolumn{1}{r|}{46} & 32 \\ \hline
\end{tabular}%
}
\end{table}

Lastly, we test the \textsc{WMSeg} and our \textsc{SpanSeg} when dealing with the largest benchmark dataset AS on Chinese to discuss the size of the model and the inference time. The statistics are presented in Table~\ref{timetab}, showing that our \textsc{SpanSeg} has the smaller size and faster inference time than \textsc{WMSeg}. The statistics can be explained by \textsc{WMSeg} \cite{tian-etal-2020-improving-chinese} containing word-hood memory networks to encode both n-grams and the word-hood information, while our \textsc{SpanSeg} encodes n-grams information via span representation.

\subsection{Analysis}

\begin{table}[H]
\caption{\label{viambi}Error statistics of the overlapping ambiguity problem involving three consecutive tokens on VWS dataset. The symbols \cmark~and \xmark~denote predicting correctly and incorrectly, respectively.}
\centering
\resizebox{0.6\textwidth}{!}{%
\begin{tabular}{c|c|r|r|r|r}
\hline
\multicolumn{1}{l|}{\multirow{2}{*}{\textbf{BiLSTM-CRF}}} & \multicolumn{1}{l|}{\multirow{2}{*}{\textbf{\textsc{SpanSeg}}}} & \multicolumn{4}{c}{\textbf{Configuration}} \\ \cline{3-6} 
\multicolumn{1}{l|}{} & \multicolumn{1}{l|}{} & \multicolumn{1}{c|}{\textbf{Defalut}} & \multicolumn{1}{c|}{\textbf{XLM-R}} & \multicolumn{1}{c|}{\textbf{TAG}} & \multicolumn{1}{c}{\textbf{TAG+XLM-R}} \\ \hline
\xmark & \xmark & 15 & 5 & 19 & 7 \\ \hline
\cmark & \xmark & \textbf{7} & \textbf{0} & 4 & 0 \\ \hline
\xmark & \cmark & \textbf{7} & \textbf{0} & \textbf{18} & \textbf{1} \\ \hline
\end{tabular}%
}
\end{table}

To explore how our \textsc{SpanSeg} learns to predict VWS and CWS, we select the statistics of the overlapping ambiguity problem involving three consecutive tokens. The first case is that given the gold standard tags ``\textbf{\color{blue}\underline{B E} S}'', the prediction is incorrect if its tags ``\textbf{\color{red}S \underline{B E}}'', and is correct if its tags ``\textbf{\color{blue}\underline{B E} S}''. The second case is that given the gold standard tags ``\textbf{\color{blue}S \underline{B E}}'', the prediction is incorrect if its tags ``\textbf{\color{red}\underline{B E} S}'', and is correct if its tags ``\textbf{\color{blue}S \underline{B E}}''. Notably, we do not count the case that is not one in two cases we describe. We present the error statistics for Vietnamese in Table~\ref{viambi}. We can see that the contextual information from XLM-R helps both BiLSTM-CRF and our \textsc{SpanSeg} in reducing ambiguity. However, according to Table~\ref{mainvtb}, the predicted word boundary information helps both BiLSTM-CRF and our \textsc{SpanSeg} in increasing overall performance but causes the overlapping ambiguity problem. Our \textsc{SpanSeg} (TAG) solves overlapping ambiguity better than BiLSTM-CRF (TAG) when utilizing predicted word boundary information. Lastly, we also provide error statistics for Chinese in Table~\ref{zhambi}. We can see that overlapping ambiguity is the crucial problem for both \textsc{WMSeg} \cite{tian-etal-2020-improving-chinese} and our \textsc{SpanSeg} on MSR, PKU, and AS datasets.

\begin{table}[H]
\caption{\label{zhambi}Error statistics of the overlapping ambiguity problem involving three consecutive tokens on five Chinese benchmark datasets. The symbols \cmark~and \xmark~denote predicting correctly and incorrectly, respectively.}
\centering
\resizebox{0.6\textwidth}{!}{%
\begin{tabular}{c|c|r|r|r|r|r}
\hline
\textbf{\textsc{\textbf{WMSeg}}} \cite{tian-etal-2020-improving-chinese} & \textbf{\textsc{\textbf{SpanSeg}}} & \multicolumn{1}{c|}{\textbf{MSR}} & \multicolumn{1}{c|}{\textbf{PKU}} & \multicolumn{1}{c|}{\textbf{AS}} & \multicolumn{1}{c|}{\textbf{\textsc{CityU}}} & \multicolumn{1}{c}{\textbf{CTB6}} \\ \hline
\xmark & \xmark & 14 & 13 & 12 & 2 & 3 \\ \hline
\cmark & \xmark & \textbf{2} & \textbf{2} & 2 & \textbf{1} & \textbf{2} \\ \hline
\xmark & \cmark & \textbf{2} & 1 & \textbf{5} & 0 & 0 \\ \hline
\end{tabular}%
}
\end{table}

\section{Conclusion}
This paper proposes a span labeling approach, namely \textsc{SpanSeg}, for VWS. Straightforwardly, our approach encodes the n-gram information by using span representations. We evaluate our \textsc{SpanSeg} on the Vietnamese treebank dataset for the word segmentation task with the predicted word boundary information and the contextual pre-trained embedding from the XLM-RoBerta model. The experimental results on VWS show that our \textsc{SpanSeg} is better than BiLSTM-CRF when utilizing the predicted word boundary and contextual information with the state-of-the-art F-score of 98.31\%. We also evaluate our \textsc{SpanSeg} on five Chinese benchmark datasets to verify our approach. Our \textsc{SpanSeg} achieves competitive or higher F-scores through experimental results, fewer parameters, and faster inference time than the previous state-of-the-art method, \textsc{WMSeg}. Lastly, we also show that overlapping ambiguity is a complex problem for VWS and CWS. Via the error analysis on the Vietnamese treebank dataset, we found that utilizing the predicted word boundary information causes overlapping ambiguity; however, our \textsc{SpanSeg} is better than BiLSTM-CRF in this case. Finally, our \textsc{SpanSeg} will be made available to the open-source community for further research and development.

\bibliographystyle{splncs04nat}
\bibliography{main}

\begin{thebibliography}{34}
\providecommand{\natexlab}[1]{#1}
\providecommand{\url}[1]{\texttt{#1}}
\providecommand{\urlprefix}{URL }
\expandafter\ifx\csname urlstyle\endcsname\relax
  \providecommand{\doi}[1]{doi:\discretionary{}{}{}#1}\else
  \providecommand{\doi}{doi:\discretionary{}{}{}\begingroup
  \urlstyle{rm}\Url}\fi

\bibitem[{Chen et~al.(2015)Chen, Qiu, Zhu, Liu, and
  Huang}]{chen-etal-2015-long}
Chen, X., Qiu, X., Zhu, C., Liu, P., Huang, X.: {Long Short-Term Memory Neural
  Networks for {C}hinese Word Segmentation}. In: {Proceedings of EMNLP}, pp.
  1197--1206 (2015)

\bibitem[{Chen et~al.(2017)Chen, Shi, Qiu, and
  Huang}]{chen-etal-2017-adversarial}
Chen, X., Shi, Z., Qiu, X., Huang, X.: {Adversarial Multi-Criteria Learning for
  {C}hinese Word Segmentation}. In: {Proceedings of ACL}, pp. 1193--1203 (2017)

\bibitem[{Conneau et~al.(2020)Conneau, Khandelwal, Goyal, Chaudhary, Wenzek,
  Guzm{\'a}n, Grave, Ott, Zettlemoyer, and
  Stoyanov}]{conneau-etal-2020-unsupervised}
Conneau, A., Khandelwal, K., Goyal, N., Chaudhary, V., Wenzek, G., Guzm{\'a}n,
  F., Grave, E., Ott, M., Zettlemoyer, L., Stoyanov, V.: {Unsupervised
  Cross-lingual Representation Learning at Scale}. In: {Proceedings of ACL},
  pp. 8440--8451 (2020)

\bibitem[{Devlin et~al.(2019)Devlin, Chang, Lee, and
  Toutanova}]{devlin-etal-2019-bert}
Devlin, J., Chang, M.W., Lee, K., Toutanova, K.: {{BERT}: Pre-training of Deep
  Bidirectional Transformers for Language Understanding}. In: {Proceedings of
  NAACL}, pp. 4171--4186 (2019)

\bibitem[{Diao et~al.(2020)Diao, Bai, Song, Zhang, and
  Wang}]{diao-etal-2020-zen}
Diao, S., Bai, J., Song, Y., Zhang, T., Wang, Y.: {{ZEN}: Pre-training
  {C}hinese Text Encoder Enhanced by N-gram Representations}. In: {Findings of
  EMNLP}, pp. 4729--4740 (2020)

\bibitem[{Dinh et~al.(2001)Dinh, Hoang, and Nguyen}]{diend1}
Dinh, D., Hoang, K., Nguyen, V.T.: {Vietnamese Word Segmentation}. In:
  {Proceedings of the Sixth Natural Language Processing Pacific Rim Symposium},
  pp. 749--756 (2001)

\bibitem[{Dozat and Manning(2017)}]{dozat2017deep}
Dozat, T., Manning, C.D.: {Deep Biaffine Attention for Neural Dependency
  Parsing}. In: {Proceedings of ICLR} (2017)

\bibitem[{Emerson(2005)}]{emerson-2005-second}
Emerson, T.: {The Second International {C}hinese Word Segmentation Bakeoff}.
  In: {Proceedings of the Fourth {SIGHAN} Workshop on {C}hinese Language
  Processing} (2005)

\bibitem[{Gong et~al.(2019)Gong, Chen, Gui, and Qiu}]{GongCGQ19}
Gong, J., Chen, X., Gui, T., Qiu, X.: {Switch-LSTMs for Multi-Criteria Chinese
  Word Segmentation}. In: {Proceedings of AAAI}, pp. 6457--6464 (2019)

\bibitem[{Higashiyama et~al.(2019)Higashiyama, Utiyama, Sumita, Ideuchi, Oida,
  Sakamoto, and Okada}]{higashiyama-etal-2019-incorporating}
Higashiyama, S., Utiyama, M., Sumita, E., Ideuchi, M., Oida, Y., Sakamoto, Y.,
  Okada, I.: {Incorporating Word Attention into Character-Based Word
  Segmentation}. In: {Proceedings of NAACL}, pp. 2699--2709 (2019)

\bibitem[{{Ilya Loshchilov and Frank Hutter}(2019)}]{loshchilov2019decoupled}
{Ilya Loshchilov and Frank Hutter}: {Decoupled Weight Decay Regularization}.
  In: {Proceedings of ICLR} (2019)

\bibitem[{Ke et~al.(2021)Ke, Shi, Sun, Meng, Wang, and Qiu}]{ke-etal-2021-pre}
Ke, Z., Shi, L., Sun, S., Meng, E., Wang, B., Qiu, X.: {Pre-training with Meta
  Learning for {C}hinese Word Segmentation}. In: {Proceedings of NAACL}, pp.
  5514--5523 (2021)

\bibitem[{Le et~al.(2008)Le, Nguyen, Roussanaly, and Ho}]{hongphuong08}
Le, H.P., Nguyen, T.M.H., Roussanaly, A., Ho, T.V.: {A Hybrid Approach to Word
  Segmentation of Vietnamese Texts}. In: {Language and Automata Theory and
  Applications}, pp. 240--249, {Springer Berlin Heidelberg} (2008)

\bibitem[{{Luu} and {Yamamoto}(2012)}]{luu2012}
{Luu}, T.A., {Yamamoto}, K.: \begin{vi}Ứng dụng phương pháp {P}ointwise
  vào bài toán tách từ cho tiếng {V}iệt\end{vi} (2012),
  \urlprefix\url{http://www.vietlex.com/xu-li-ngon-ngu/117-Ung_dung_phuong_phap_Pointwise_vao_bai_toan_tach_tu_cho_tieng_Viet}

\bibitem[{Ma et~al.(2018)Ma, Ganchev, and Weiss}]{ma-etal-2018-state}
Ma, J., Ganchev, K., Weiss, D.: {State-of-the-art {C}hinese Word Segmentation
  with {B}i-{LSTM}s}. In: {Proceedings of EMNLP}, pp. 4902--4908 (2018)

\bibitem[{Nguyen et~al.(2006)Nguyen, Nguyen, Phan, Nguyen, and
  Ha}]{nguyenetal2006}
Nguyen, C.T., Nguyen, T.K., Phan, X.H., Nguyen, L.M., Ha, Q.T.: {{V}ietnamese
  Word Segmentation with {CRF}s and {SVM}s: An Investigation}. In: {Proceedings
  of PACLIC}, pp. 215--222, {Tsinghua University Press} (2006)

\bibitem[{Nguyen(2019)}]{nguyen-2019-neural}
Nguyen, D.Q.: {A neural joint model for {V}ietnamese word segmentation, {POS}
  tagging and dependency parsing}. In: {Proceedings of the The 17th Annual
  Workshop of the Australasian Language Technology Association}, pp. 28--34
  (2019)

\bibitem[{Nguyen et~al.(2018)Nguyen, Nguyen, Vu, Dras, and Johnson}]{datnq1}
Nguyen, D.Q., Nguyen, D.Q., Vu, T., Dras, M., Johnson, M.: {A Fast and Accurate
  Vietnamese Word Segmenter}. In: {Proceedings of LREC}, pp. 2582--2587 (2018)

\bibitem[{Nguyen et~al.(2017)Nguyen, Vu, Nguyen, Dras, and
  Johnson}]{nguyen-etal-2017-word}
Nguyen, D.Q., Vu, T., Nguyen, D.Q., Dras, M., Johnson, M.: {From Word
  Segmentation to {POS} Tagging for {V}ietnamese}. In: {Proceedings of the
  Australasian Language Technology Association Workshop 2017}, pp. 108--113
  (2017)

\bibitem[{Nguyen et~al.(2019)Nguyen, Van~Thin, Van~Nguyen, and
  Nguyen}]{nguyen-etal-2019-ws-svm}
Nguyen, D.V., Van~Thin, D., Van~Nguyen, K., Nguyen, N.L.T.: {Vietnamese Word
  Segmentation with SVM: Ambiguity Reduction and Suffix Capture}. In:
  {Proceedings of PACLING}, pp. 400--413 (2019)

\bibitem[{Nguyen et~al.(2009)Nguyen, Vu, Nguyen, Nguyen, and
  Le}]{nguyen-etal-2009-building}
Nguyen, P.T., Vu, X.L., Nguyen, T.M.H., Nguyen, V.H., Le, H.P.: {Building a
  Large Syntactically-Annotated Corpus of {V}ietnamese}. In: {Proceedings of
  the Third Linguistic Annotation Workshop ({LAW} {III})}, pp. 182--185 (2009)

\bibitem[{Nguyen and Le(2016)}]{phongnt1}
Nguyen, T.P., Le, A.C.: {A hybrid approach to Vietnamese word segmentation}.
  In: {Proceeding of IEEE-RIVF}, pp. 114--119 (2016)

\bibitem[{Qiu et~al.(2020)Qiu, Pei, Yan, and Huang}]{qiu-etal-2020-concise}
Qiu, X., Pei, H., Yan, H., Huang, X.: {A Concise Model for Multi-Criteria
  {C}hinese Word Segmentation with Transformer Encoder}. In: {Findings of
  EMNLP}, pp. 2887--2897 (2020)

\bibitem[{Song et~al.(2006)Song, Guo, and Cai}]{song-etal-2006-chinese}
Song, Y., Guo, J., Cai, D.: {{C}hinese Word Segmentation Based on an Approach
  of Maximum Entropy Modeling}. In: {Proceedings of the Fifth {SIGHAN} Workshop
  on {C}hinese Language Processing}, pp. 201--204 (2006)

\bibitem[{Stern et~al.(2017)Stern, Andreas, and
  Klein}]{stern-etal-2017-minimal}
Stern, M., Andreas, J., Klein, D.: {A Minimal Span-Based Neural Constituency
  Parser}. In: {Proceedings of ACL}, pp. 818--827 (2017)

\bibitem[{Sun et~al.(1998)Sun, Shen, and Tsou}]{sun-etal-1998-chinese}
Sun, M., Shen, D., Tsou, B.K.: {{C}hinese Word Segmentation without Using
  Lexicon and Hand-crafted Training Data}. In: {Proceedings of ACL-COLING}, pp.
  1265--1271 (1998)

\bibitem[{Tian et~al.(2020)Tian, Song, Xia, Zhang, and
  Wang}]{tian-etal-2020-improving-chinese}
Tian, Y., Song, Y., Xia, F., Zhang, T., Wang, Y.: {Improving {C}hinese Word
  Segmentation with Wordhood Memory Networks}. In: {Proceedings of ACL}, pp.
  8274--8285 (2020)

\bibitem[{Tseng et~al.(2005)Tseng, Chang, Andrew, Jurafsky, and
  Manning}]{tseng-etal-2005-conditional}
Tseng, H., Chang, P., Andrew, G., Jurafsky, D., Manning, C.: {A Conditional
  Random Field Word Segmenter for Sighan Bakeoff 2005}. In: {Proceedings of the
  Fourth {SIGHAN} Workshop on {C}hinese Language Processing} (2005)

\bibitem[{Wang and Xu(2017)}]{wang-xu-2017-convolutional}
Wang, C., Xu, B.: {Convolutional Neural Network with Word Embeddings for
  {C}hinese Word Segmentation}. In: {Proceedings of IJCNLP}, pp. 163--172
  (2017)

\bibitem[{Xu and Sun(2016)}]{xu-sun-2016-dependency}
Xu, J., Sun, X.: {Dependency-based Gated Recursive Neural Network for {C}hinese
  Word Segmentation}. In: Proceedings of ACL), pp. 567--572 (2016)

\bibitem[{Xue et~al.(2005)Xue, Xia, Chiou, and
  Palmer}]{xue_xia_chiou_palmer_2005}
Xue, N., Xia, F., Chiou, F.D., Palmer, M.: {The Penn Chinese TreeBank: Phrase
  structure annotation of a large corpus}. {Natural Language Engineering}
  \textbf{11}(2), 207–238 (2005)

\bibitem[{Zhang et~al.(2016)Zhang, Zhang, and
  Fu}]{zhang-etal-2016-transition-based}
Zhang, M., Zhang, Y., Fu, G.: {Transition-Based Neural Word Segmentation}. In:
  {Proceedings of ACL}, pp. 421--431 (2016)

\bibitem[{Zhang et~al.(2020)Zhang, Zhou, and Li}]{ijcai2020-560}
Zhang, Y., Zhou, H., Li, Z.: {Fast and Accurate Neural CRF Constituency
  Parsing}. In: {Proceedings of IJCAI}, pp. 4046--4053 (2020)

\bibitem[{Zhou et~al.(2017)Zhou, Yu, Zhang, Huang, Dai, and
  Chen}]{zhou-etal-2017-word}
Zhou, H., Yu, Z., Zhang, Y., Huang, S., Dai, X., Chen, J.: {Word-Context
  Character Embeddings for {C}hinese Word Segmentation}. In: {Proceedings of
  EMNLP}, pp. 760--766 (2017)

\end{thebibliography}

\end{document}